# Real-Time and Robust Method for Hand Gesture Recognition System Based on Cross-Correlation Coefficient

Reza Azad[1], Babak Azad [2] and Iman tavakoli kazerooni [3]

[1] IEEE Member, Electrical and Computer Engineering Department, Shahid Rajaee Teacher training University
Tehran, Iran
rezazad68@gmail.com

[2] Institute of Computer science, Shahid Bahonar University
Shiraz, Iran
babak.babi72@gmail.com

[3] Faculty of Computer science, Hamedan science and research University
Hamedan, Iran
iman_tavakoli2008@yahoo.com

**Abstract**
Hand gesture recognition possesses extensive applications in virtual reality, sign language recognition, and computer games. The direct interface of hand gestures provides us a new way for communicating with the virtual environment. In this paper a novel and real-time approach for hand gesture recognition system is presented. In the suggested method, first, the hand gesture is extracted from the main image by the image segmentation and morphological operation and then is sent to feature extraction stage. In feature extraction stage the Cross-correlation coefficient is applied on the gesture to recognize it. In the result part, the proposed approach is applied on American Sign Language (ASL) database and the accuracy rate obtained 98.34%.
***Keywords:*** *Hand Gesture Recognition, Image Processing, Virtual Reality, Cross-correlation.*

## 1. Introduction

Hand gesture recognition possesses extensive applications in sign language recognition, computer games [1] and virtual reality [2-4]. A gesture is a form of non-verbal communication in which visible bodily actions can be used for communication. It can be categorized into static and dynamic [5]. The process of recognizing and predicting a gesture is known as Gesture Recognition and Sign language recognition is one of its applications. Sign language can involve combining orientation and movements of the hands, arms or body, hand shapes, and facial expressions to express thoughts and words which can be used for communication mostly by Deaf and Dumb people. It can also provide a good interface between computer and user, so in this paper we are representing a hand gesture recognition system which can recognize most of the character from ASL with a good accuracy [6]. Figure 1 shows the ASL symbols. In some passed decades Gesture recognition becomes very influencing term. There were many gesture recognition techniques developed for tracking and recognizing various hand gestures. Each one of them has their pros and cons. The older one is wired technology, in which users need to tie up themselves with the help of wire in order to connect or interface with the computer system. In wired technology user cannot freely move in the room as they connected with the computer system via wire and limited with the length of wire. Instrumented gloves also called electronics gloves or data gloves is the example of wired technology. These instrumented gloves made up of some sensors, provide the information related to hand location, finger position orientation etc. through the use of sensors. These data gloves provide good results but they are extremely expensive to utilize in wide range of common application.

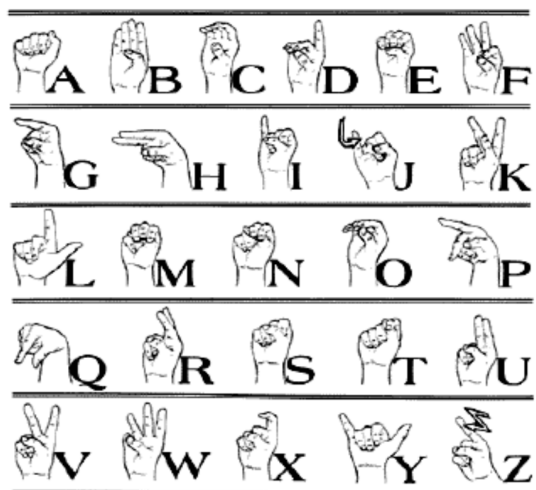

Fig. 1 American Sign Language that mentioned in [6].

Data gloves are then replaced by optical markers. These optical markers project Infra-Red light and reflect this light on screen to provide the information about the location of hand or tips of fingers wherever the markers are wear on hand, the corresponding portion will display on the screen. These systems also provide the good result but require very complex configuration. Later on some advanced techniques have been introduced like Image based techniques which requires processing of image features like texture, color etc. the methods on optical markers are very expensive and have very complex configuration. The method based on image processing is weak against under different illumination condition, color texture modifying, which leads to changes in observed results. For utilizing the image processing method for hand gestures recognition system we proposed current paper method. In [7-9] a vision based methods for recognition hand gesture is presented. These methods are based on shape features and are highly influenced by some constraints like hand and noise. Our approach for hand gesture recognition is consist of three step, I. Image segmentation; II. Morphological Filtering; III. Cross-correlation based feature extraction and matching. In section two the database is described, in section three proposed methods is presented and in section four and five the practical result and conclusion are presented respectively.

## 2. Database Description

Our approach for hand gesture recognition is based on static mode so; our first problem is to gather a good quality of data since our classifier will classify characters according to it only. We had created our own database for each character of ASL, which can includes 504 images i.e. 21 images for each (24) gestures. During creating a database images captured should have uniform dark color background that can be black with a white color rubber glove on hand as in contrast. We had done this in order to minimize noise and unwanted data so that we can easily do segmentation process. The user has to wear a black color cloth around his arm till wrist from the shoulder so that black color cloth can easily match with the background. Covered arm and the background should be of similar color. Letter j and z are discarded because we can describe them dynamically only and our approach is for static gestures only. Figure 2 shows the sample of our database that are more like a [6] databases.

## 3. Proposed Method

Our approach for hand gesture recognition is consist of three step, I. Image segmentation; II. Morphological Filtering; III. Cross-correlation based feature extraction and matching. First of all we have to do preprocessing of data which is very important task in Hand Gesture Recognition system. Preprocessing should be done on images initially before we extract features from data of Hand Gestures. It is to be done to remove noise, unwanted errors and to make data efficient so, that it can be used for further image processing. We used two steps for preprocessing of data 1.Image Segmentation 2.Morphological Filtering [6].

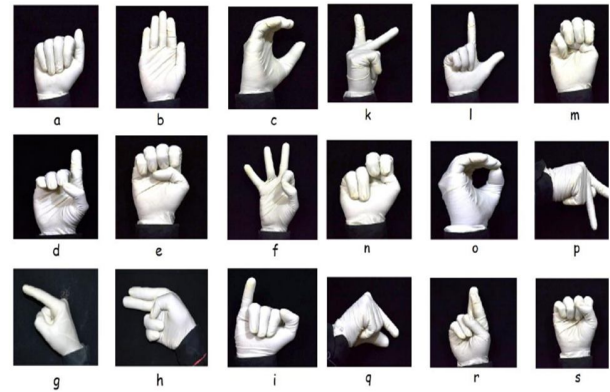

Fig. 2 Sample of images from our database

### 3.1 Image Segmentation

It is a process in which we convert a RGB image or gray scale image into binary (Black and White) image. This is to be done because we can get only two objects i.e. black and white only in our image. Black with the background and white represents our hand, Qtsu algorithm is used to convert image into binary [10]. A good segmentation process is that process in which background doesn't denote any part of hand and hand shouldn't have any part of background. To obtain best result we have to choose best possible threshold value and segmentation can be done according to that value. The selection of the segmentation technique mainly depends on the type of image on which we have to do processing and Qtsu algorithms [10] had been tested and work efficiently with our hand gestures data. It is an unsupervised and non-parametric method of segmentation which can select threshold automatically and do segmentation [9]. Suppose there are two classes of pixels with $\Omega 0$ as background pixel and $\Omega 1$ as hand pixel. $\Omega 0$ shows the pixels with intensity level $[1,2.......K]$ and $\Omega 1$ shows the pixels with intensity level $[K+1......L]$, from these classes we get the threshold value $K^*$ which is in between value of K and K+1 and now hand pixel is assigned value "1" and background pixel is assigned value "0" and we get our desired binary image [11].

## 3.2 Morphological Filtering

The segmented images we get after applying the Otsu algorithm are not perfectly processed and further processing is needed in those images to remove unwanted data and errors. There are still some background parts which contain 1s and some hand parts which denote 0s. In order to remove that noise we have to apply morphological filtering techniques on those segmented images. It is necessary to remove these errors as they can create problem in recognition of hand gestures and reduce the system efficiency [12]. So, morphological filtering is necessary to applied on segmented images and then we get a better smooth, closed and contour of a gesture. Dilation, Erosion, Opening, and Closing is the basic operators that work in morphological filtering [13]. Sample of preprocessing result is shown in Figure 3 and the experiments are performed in MATLAB. After the preprocessing we get a smooth and better hand gesture which can results a better efficiency [6].

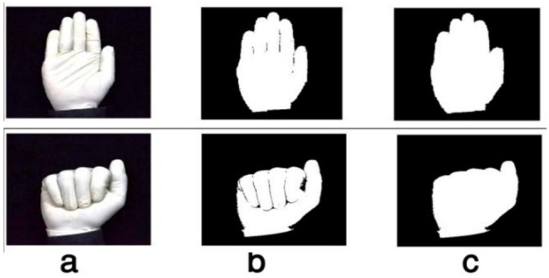

Fig. 3 Result of the gestures after image preprocessing (a): main image (b): image after image segmentation (c): after morphological operation

## 3.3 Cross-correlation Coefficient

Now we have to extract feature for gesture recognition. For feature extraction and matching we used Cross-correlation Coefficient. In signal processing, cross-correlation is a measure of similarity of two waveforms as a function of a time-lag applied to one of them. In this part we used this function for matching of hand gesture. The cross correlation coefficient is defined as Equation (1):

$$\gamma(x,y) = \frac{\sum_s \sum_t \delta_{I(x+s,y+t)} \delta_T(s,t)}{\sum_s \sum_t \delta^2_{I(x+s,y+t)} \delta^2_T(s,t)} \quad (1)$$

Where $\delta_{I(x+s,y+t)} = I(x+s,y+t) - I'(x,y)$,
$\delta_T(s,t) = T(s,t) - T'$,
$s \epsilon \{1,2,3,\ldots,p\}$,
$t \epsilon \{1,2,3,\ldots,q\}$,
$x \epsilon \{1,2,3,\ldots,m-p+1\}$,
$y \epsilon \{1,2,3,\ldots,n-q+1\}$,
$I'(x,y) = \frac{1}{pq} \sum_s \sum_t I(x+s,y+t)$

$I' = \frac{1}{pq} \sum_s \sum_t T(s,t)$

The value of cross-correlation coefficient $\gamma$ ranges from –l to +l corresponds completely not matched and completely matched respectively. For template matching the template, T slides over I and $\gamma$ is calculated for each coordinate (x, y). After calculation, the point which exhibits maximum $\gamma$ is referred to as the match point. The following step is used for matching of hand gesture:

Step 1: A hand gesture template of size m × n is taken.
Step 2: The normalized 2-D auto-correlation of hand gesture template is found out.
Step 3: The normalized 2-D cross-correlation of hand gesture template with various template is calculated.
Step 4: The mean squared error (MSE) of auto correlation and cross-correlation of different sample are found out. The minimum MSE is found out and stored.
Step 5: The corresponding minimum MSE represent the recognized gesture.

## 4. Practical Result

Our suggestive method have been done on Intel Core i3-2330M CPU, 2.20 GHz with 2 GB RAM under Matlab environment. Fig. 4 shows the face of worked systems.

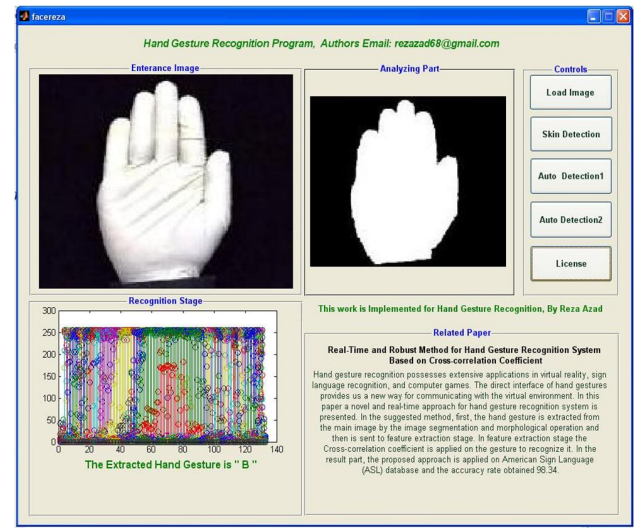

Fig. 4 Hand gesture recognition system

In this study, for experimental analysis, we had applied the above mention technique on our database of American Sign Language which consists of 504 images i.e. 21 images per character and we was able to recognize 24 characters out of 26 characters from sign language because the remaining two characters gesture are dynamic and the developed approach is for static characters only. Table 1 shows the accuracy rate for each hand gesture.

Table 1: Accuracy rate for each group of signs

| Hand Gesture | Input Image | Recognized Image | Accuracy Rate |
|---|---|---|---|
| A | 21 | 20 | 95.23% |
| B | 21 | 21 | 100% |
| C | 21 | 21 | 100% |
| D | 21 | 21 | 100% |
| E | 21 | 19 | 90.47% |
| F | 21 | 21 | 100% |
| G | 21 | 21 | 100% |
| H | 21 | 21 | 100% |
| I | 21 | 21 | 100% |
| K | 21 | 21 | 100% |
| L | 21 | 21 | 100% |
| M | 21 | 19 | 90.47% |
| N | 21 | 21 | 100% |
| O | 21 | 21 | 100% |
| P | 21 | 21 | 100% |
| Q | 21 | 21 | 100% |
| R | 21 | 21 | 100% |
| S | 21 | 20 | 95.23% |
| T | 21 | 21 | 100% |
| U | 21 | 21 | 100% |
| V | 21 | 21 | 100% |
| W | 21 | 21 | 100% |
| X | 21 | 21 | 100% |
| Y | 21 | 21 | 100% |
| Total | 504 | 498 | 98.80% |

In our experiment (with the 98.80% accuracy), we observed confusion hand gesture in the recognition phase between some signs. The major confusions were amongst A, S and E, M. This happened because A, S and E, M look like each other in some samples. Figure 5 shows the success and confuse rate for each group of signs in recognition stage.

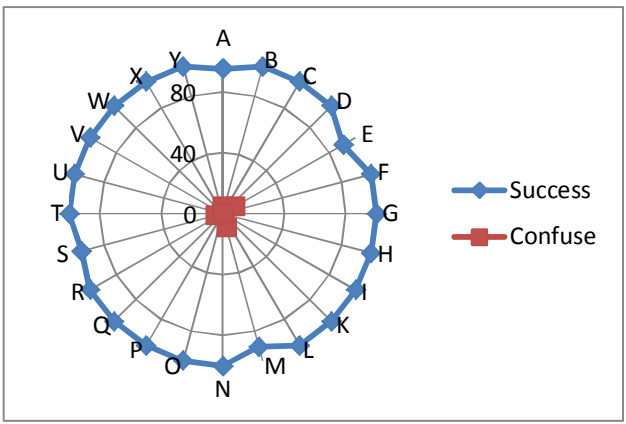

Fig. 5 Success and confuse rate for each group of signs

## 5. Conclusions

In this paper we proposed a novel and real-time approach for hand gesture recognition system. In the mentioned method, first, the hand gesture is extracted from the main image by the image segmentation and morphological operation and then is sent to feature extraction stage. In feature extraction stage the Cross-correlation coefficient is applied on the gesture to recognize it. High recognition rate showed our proposed approach ability in hand gesture recognition. In the future we have plane to recognize dynamic hand gesture recognition in video sequence for virtual reality.

## 6. Acknowledgement

This work is supported by the SRTTU under Reza Azad hand gesture recognition project. The authors would like to thank prof. Panahi (Iranian center of foreign language learning chief, Ardebil) for his help in English writing.

**Reza Azad** was born in Ardebil, Iran, in 1989. He is now studying B.Sc. in university of Shahid Rajaee Teacher Training, Tehran, Iran, from 2011 until now in computer software engineering technology. He takes the fourth place at Iranian university entering exam. Also he's a member of IEEE, member of elites of the country, top and Honor student in university. Hi have 7 papers in the international conference and 4 papers in international journal. In 2013 His two papers picked out as high level in science and innovation by the CITADIM 2013 and published in international high level scientific journal also He dominated as best researcher in 2013 by the SRTTU Computer faculty. As a Reviewer in the 3rd IEEE International Conference on Computer and Knowledge Engineering. His research interests include image processing, artificial intelligence, handwritten character recognition, skin detection, face detection and recognition, human tracking, pattern recognition, virtual reality, machine learning and localization of autonomous vehicles.

**Babak Azad** was born in Ardebil, Iran, in 1993. He is studying B.Sc. in University of Shahid Bahonar, Shiraz, in 2013 in computer software engineering and he is top student in university. Hi have one paper in the international conference, one paper in international journal and 3 papers in regional conference. His research interests include image processing, pattern recognition, cloud computing, information security.

**Iman tavakoli kazerooni** was born in kazeroon, Iran, in 1987. He completed his undergraduate educations in 2001. Then he passed the bachelor education in computer science at Bahonar University of Shiraz. He currently studies his master in Hamadan science and Research University until now. Also hi is an educator in mamasani azad bahonar university, minab azad university and darab university faculty His favorite research fields are: computer vision, web designing and programming.